\documentclass{article}

\usepackage{spconf}
\usepackage{amsmath}
\usepackage{graphicx}
\usepackage{bm}
\usepackage{amssymb}
\usepackage{caption}
\usepackage{subcaption}
\usepackage{placeins}
\usepackage{algorithm}
\usepackage[noend]{algpseudocode}
\usepackage{mathtools}
\usepackage{multirow}

\newcommand{\R}{\mathbb{R}}

\newcommand{\vect}{\mathbf{vec}}
\newcommand{\OptShrink}{\mathbf{OptShrink}}
\newcommand{\soft}{\mathbf{soft}}
\newcommand{\TV}{\mathbf{TV}}
\newcommand{\TVDN}{\mathbf{TVDN}}

\newcommand{\sign}{\mathbf{sign}}

\newcommand{\ie}{\textit{i}.\textit{e}.}

\newcommand{\proj}{\mathcal{P}}

\title{AUGMENTED ROBUST PCA FOR FOREGROUND-BACKGROUND SEPARATION\\
ON NOISY, MOVING CAMERA VIDEO}
\name{Chen Gao, Brian E. Moore, and Raj Rao Nadakuditi
\thanks{This work was supported in part by the following grants: ONR grant N00014-15-1-2141, DARPA Young Faculty Award D14AP00086, and ARO MURI grants W911NF-11-1-0391 and 2015-05174-05.}}
\address{Department of EECS, University of Michigan, Ann Arbor, MI, 48109}

\begin{document}
\ninept

\maketitle

\begin{abstract}
This work presents a novel approach for robust PCA with total variation regularization for foreground-background separation and denoising on noisy, moving camera video. Our proposed algorithm registers the raw (possibly corrupted) frames of a video and then jointly processes the registered frames to produce a decomposition of the scene into a low-rank background component that captures the static components of the scene, a smooth foreground component that captures the dynamic components of the scene, and a sparse component that isolates corruptions. Unlike existing methods, our proposed algorithm produces a panoramic low-rank component that spans the entire field of view, automatically stitching together corrupted data from partially overlapping scenes. The low-rank portion of our robust PCA model is based on a recently discovered optimal low-rank matrix estimator (OptShrink) that requires no parameter tuning. We demonstrate the performance of our algorithm on both static and moving camera videos corrupted by noise and outliers.
\end{abstract}


\vspace{-1mm}
\section{Introduction} \label{sec:intro}
Video processing methods are an important class of algorithms in computer vision because video data is a rich source of semantic information. In this work, we focus on the problem of robust foreground-background separation, where one seeks to decompose a scene into a static background and dynamic foreground in the presence of noise or other corruptions. Decompositions of this form are useful because the constituent components play important roles in various computer vision problems, such as motion detection~\cite{huang2011advanced}, object recognition~\cite{tsaig2002automatic}, moving object detection~\cite{bouwmans2014robust}, ~\cite{sobral2014comprehensive} and video coding~\cite{ye2015foreground}, especially when there are active, moving objects of interest and a relatively static background~\cite{he2012incremental}. For example, in background subtraction~\cite{elgammal2000non}, one discriminates moving objects from their static background by subtracting a reference background model from the current frame.

There has been much recent work on the foreground-background separation problem. A prominent method is robust principal component analysis (RPCA) \cite{candes2011robust,guyon2012foreground,zhou2013shifted}, which uses a low-rank subspace model to estimate the background and a spatially sparse model to estimate the foreground. Recent work has extended this model to the case of robust tensor decompositions \cite{zhang2014novel}. Alternatively, supervised approaches like GMM ~\cite{stauffer1999adaptive} learn a model of the background from labeled training data. Recently, a TVRPCA~\cite{cao2016total} method was proposed to separate dynamic background from moving objects using total variation (TV)-based regularization to model the spatial continuity of the foreground. In the non-static background case, an RPCA-based model was proposed in \cite{ebadi2016approximated} that iteratively estimates the decomposition along with the parameters of an affine transformation model that describes the motion of the frames; however, the approach considers only the intersection (common view) of the video. Another approach is DECOLOR \cite{zhou2013moving}, which employs $\ell_0$-regularized RPCA and a Markov random field (MRF) model to iteratively decompose the scene into foreground and background.

In this paper, we propose a robust foreground-background separation and denoising algorithm that can decompose a noisy, moving camera video into a panoramic low-rank background component that spans the entire field of view and a smooth foreground component. Our algorithm proceeds by registering the frames of the raw video to a common reference perspective and then minimizing a modified robust PCA cost that accounts for the partially overlapping views of registered frames and includes TV regularization to decouple the foreground from noise and sparse corruptions.
	
The paper is organized as follows. In Section~\ref{sec:Registration}, we describe our video registration strategy. Section~\ref{sec:RPCA} formulates our augmented robust PCA algorithm, and Section~\ref{sec:result} provides experimental results that demonstrate the performance of our algorithm on corrupted static and moving camera data.

\vspace{-1mm}
\section{Video Registration} \label{sec:Registration}
The vast majority of video data gathered today is captured by moving (e.g., handheld) cameras. As such, it is necessary to \emph{register} the raw video---i.e., convert it into a common coordinate system---before the frames of the video can be jointly processed. In this work, we adopt the standard perspective projection model \cite{forsyth2002computer}, in which we relate different views of the same scene via homographic transformations.

\begin{figure*}[htb]
\centering
\centerline{\includegraphics[width=1\linewidth]{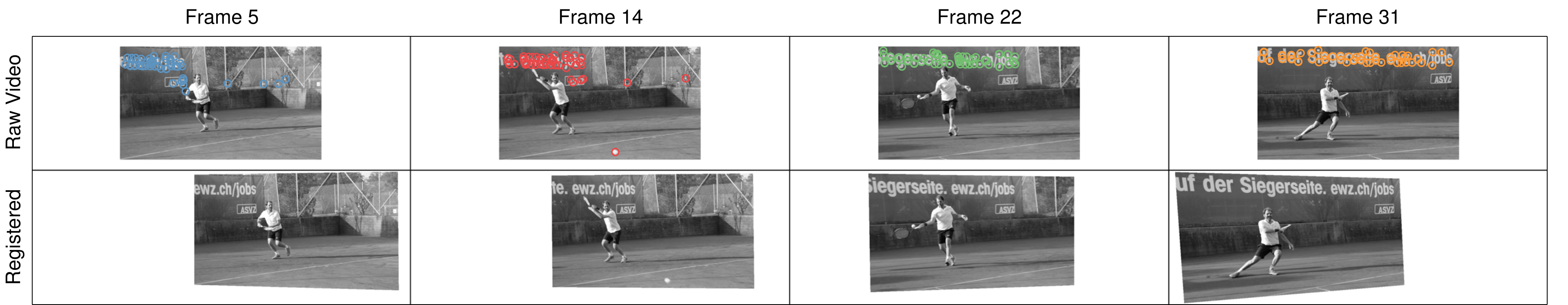}}
\caption{The video registration process. The top row depicts raw video frames $F_k$ with SURF features annotated. The bottom row depicts the corresponding registered frames $\widetilde{F}_k$ computed via \eqref{eq:registration}. The $k$-th column of the mask matrix $M \in \{0,1\}^{mn \times p}$ encodes the support of $\widetilde{F}_k$ within the aggregate view; i.e., $M_{ik} = 0$ for unobserved pixels, which are represented by white space in the registered frames above.}
\label{fig:frames}
\end{figure*}

\subsection{Registering two frames} \label{subsec:register_frames}
Consider a point $(x,y)$ in a frame that is known to correspond to a point $(\widetilde{x},\widetilde{y})$ in another frame. Under a planar surface model, one can relate the points via a projective transformation of the form
\begin{equation} \label{eq:projective}
\kappa\widetilde{p} = H^Tp,
\end{equation}
where $\widetilde{p} = [\widetilde{x}, \widetilde{y}, 1]^T$, $p = [x, y, 1]^T$, $\kappa \neq 0$ is an arbitrary scaling constant, and $H \in \mathbb{R}^{3 \times 3}$ with $H_{33} = 1$ is the transformation matrix that we would like to estimate. Given $d > 3$ correspondences $\{(x_i, y_i) \mapsto (\widetilde{x}_i, \widetilde{y}_i)\}_{i=1}^d$, one can estimate $H$ in a least squares sense by minimizing \cite{forsyth2002computer}
\begin{equation} \label{eq:projective_cost}
\min_h \|Ah\|^2 \text{~~subject to~~} h_9 = 1,
\end{equation}
where $h = \vect(H)$, $A^T = \begin{bmatrix} A_1^T, \ldots , A_d^T \end{bmatrix}$, and
\begin{equation}
A_i = \begin{bmatrix}
0 & p_i^T & -\widetilde{y}_i p_i^T\\
p_i^T & 0 & -\widetilde{x}_i p_i^T
\end{bmatrix} \in \mathbb{R}^{2 \times 9}.
\end{equation}
The solution to \eqref{eq:projective_cost} is the (scaled) smallest right singular vector of $A$.

Of course, to estimate $H$ in practice one must also solve the \emph{correspondence problem} of identifying pairs of candidate correspondences $(x_i, y_i) \mapsto (\widetilde{x}_i, \widetilde{y}_i)$ between the video frames. In this work, we adopt the standard procedure \cite{forsyth2002computer} of computing Speeded-Up Robust Features (SURF)~\cite{bay2006surf} for each frame and then using the Random Sample Consensus (RANSAC)~\cite{fischler1981random} algorithm to find a robust subset of correspondences from among the candidate features that produce a solution $\hat{H}$ to \eqref{eq:projective_cost} with small cost.

\subsection{Registering a video} \label{subsec:register_video}
One can readily extend the two-frame registration procedure from Section~\ref{subsec:register_frames} to a video by iteratively constructing homographies $H_k := H_{k \mapsto k+1}$ between frames $k$ and $k + 1$ of the video and then chaining the homographies together to map all $p$ frames to a common reference perspective (e.g., the middle frame, $\widetilde{k} = \lfloor p/2 \rfloor$). Since consecutive frames of a video are highly correlated, the homographies $H_k$ can be computed with high accuracy.

Indeed, let $F_1, \dots, F_p \in \mathbb{R}^{a \times b}$ denote the frames of a moving camera video, and denote by $\mathcal{H}_k \coloneqq \mathcal{H}_{k \mapsto k+1}$ the linear transformation that applies the projective transformation \eqref{eq:projective} defined by $H_k$ to each pixel of $F_k$. One can register the frames of the video against an anchor frame $\widetilde{k}$ by computing for each $k = 1,\ldots,p$,
\begin{equation} \label{eq:registration}
\widetilde{F}_k =
\begin{cases}
(\mathcal{H}_{\widetilde{k} - 1} \circ  \mathcal{H}_{\widetilde{k} - 2} \circ \dots \circ \mathcal{H}_k)(F_k) & k < \widetilde{k}, \\
\hphantom{(} F_k & k = \widetilde{k}, \\
(\mathcal{H}_{\widetilde{k}}^{-1} \circ  \mathcal{H}_{\widetilde{k} + 1}^{-1} \circ \dots \circ \mathcal{H}_{k - 1}^{-1})(F_k) & k > \widetilde{k}.   
\end{cases}
\end{equation}
The above procedure yields $\widetilde{F}_1,\ldots,\widetilde{F}_p \in \mathbb{R}^{m \times n}$, a collection of registered frames in a common perspective, where $m$ and $n$ are the height and width of the region defined by the union of the registered frame extents. See Figure~\ref{fig:frames} for a graphical depiction.

\section{Augmented Robust PCA algorithm} \label{sec:RPCA}
In this section, we describe our augmented robust PCA algorithm for noisy, moving camera video. Given the registered frames $\widetilde{F}_1,\ldots,\widetilde{F}_p \in \mathbb{R}^{m \times n}$ of a moving camera video, we construct the matrix $Y \in \mathbb{R}^{mn \times p}$
\begin{equation} \label{eq:Ymatrix}
Y =  \begin{bmatrix}
\vect(\widetilde{F}_1) \dots \vect(\widetilde{F}_p)
\end{bmatrix},
\end{equation}
whose columns are the vectorized registered frames. Associated with $Y$, we also define the mask matrix $M \in \{0, 1\}^{mn \times p}$ whose columns encode the support of the registered frames in the aggregate (common) view extent (see Figure \ref{fig:frames}).

The representation \eqref{eq:Ymatrix} is useful because each row of $Y$ corresponds to a fixed point in space, so we can readily apply standard static-camera models for foreground-background separation. In particular, in this work, we model the observed data $Y$ using the following (approximate) structured low-rank plus sparse model
\begin{equation} \label{eq:model}
\proj_{M}(Y) \approx \proj_{M}(L + S_1 + S_2),
\end{equation}
where $\proj_{M}$ denotes the orthogonal projection onto $M$, defined as
\begin{equation}
[\proj_{M}(X)]_{ij} = \begin{cases}
X_{ij} & M_{ij} = 1 \\
0 & M_{ij} = 0.
\end{cases}
\end{equation}
In \eqref{eq:model}, the $L$ component represents the (static) background, which we model as low-rank; $S_1$ represents sparse corruptions, which we model as a sparse matrix; $S_2$ is the foreground, which we model as a smoothly-varying matrix; and we use $\approx$ to allow for additional dense corruptions. To learn a decomposition of the form \eqref{eq:model}, we propose to solve the augmented robust PCA problem
\begin{align} \label{eq:cost}
\min_{L,S1,S2} & ~~ \frac{1}{2} \|\proj_{M}(Y - L - S_1 - S_2)\|_F^2 \ + \nonumber \\
& ~~ \lambda_L \|L\|_{\star} \ + \ \lambda_{S_1}\|S_1\|_1 \ + \ \lambda_{S_2} \TV(S_2).
\end{align}
Here, $\|.\|_*$ denotes the nuclear norm (sum of singular values), $\|.\|_1$ denotes the element-wise $\ell_1$ norm, and $\TV(.)$ denotes the total variation (TV) regularizer, a popular approach for reconstructing an image from noisy observations~\cite{chan2011augmented}. In particular, in this work, given a matrix $X \in \R^{mn \times p}$ whose columns contain the vectorized $m \times n$ spatial frames, we use the \emph{weighted} anisotropic TV of $X$:
\begin{align} \label{eq:TV}
\TV(X) := ~ \sum_{ijk} \big(& w_{ijk}^x|x_{i+1jk} - x_{ijk}| \ + \ w_{ijk}^y|x_{ij+1k}-x_{ijk}| \ + \nonumber \\[-4pt]
& w_{ijk}^z|x_{ijk+1}-x_{ijk}| \big),
\end{align}
where $x = \vect(X)$ and, with slight abuse of notation, we use $x_{ijk}$ to denote the pixel $(i,j)$ from frame $k$---i.e., the $(i + m(j - 1),k)$ entry of $X$. Here, $w_{ijk} \in \{0,1\}$ are (fixed) indicator variables that omit first differences involving unobserved pixels, i.e., those that lie outside the extent of the registered frames. The $w_{ijk}$ can be readily computed from mask $M$ (see Figure \ref{fig:frames}).


\subsection{Minimization strategy} \label{subsec:minimization}
One can solve \eqref{eq:cost} iteratively using the proximal gradient method \cite{parikh2014proximal}, for which the $L$ updates would involve applications of singular value thresholding (SVT) \cite{cai2010singular}. However, motivated by recent work \cite{moore2014improved}, we consider a modified $L$ update based on an improved low-rank matrix estimator (OptShrink) \cite{nadakuditi2014optshrink}, which has been shown to produce superior low-rank components in practice. Our proposed (modified) proximal gradient scheme thus becomes
\begin{equation} \label{eq:rpca_apgupdates_optshrink}
\begin{aligned} 
  U^{k+1} &:= \proj_{M}(L^{k+1} + S_1^{k+1} + S_2^{k+1} - Y) \\
  L^{k+1} &:= \OptShrink_r\left(L^k - \tau^k U^{k+1}\right) \\
S_1^{k+1} &:= \soft_{\tau^k\lambda_{S_1}}\left(S_1^k - \tau^k U^{k+1}\right) \\
S_2^{k+1} &:= \TVDN_{\tau^k\lambda_{S_2}}\left(S_2^k - \tau^k U^{k+1}\right),
\end{aligned} 
\end{equation}
where $\tau^k$ denotes the step size at the $k$-th iteration. In \eqref{eq:rpca_apgupdates_optshrink}, $\OptShrink(.)$ is the low-rank matrix estimator, defined for a given $r > 0$ as\footnote{See equation (14) of \cite{moore2014improved} and Algorithm~1 of \cite{nadakuditi2014optshrink} and the surrounding text for the full description and intuition behind OptShrink.}
\begin{equation} \label{eq:OptShrink}
\OptShrink_r(Z) = \sum_{i=1}^{r}\left( -2\frac{D_{\mu_Z}(\sigma_i)}{D_{\mu_Z}^\prime(\sigma_i)}\right)u_i v_i^H,
\end{equation}
where $Z = U\Sigma V^H$ is the SVD of $Z$. The OptShrink estimator computes the rank $r$ truncated SVD of its input and then applies a particular data-driven \emph{shrinkage} to the leading singular values. See \cite{nadakuditi2014optshrink} for more details. Note that, since our data $Y$ is registered, we can readily model the video background as static, in which case the low-rank component $L$ would be a rank-$1$ matrix whose columns are repeated (up to scaling) vectorized copies of the static background image. Thus, the universal parameter $r = 1$ is a natural choice, and we have essentially eliminated a tuning parameter from our model compared to the SVT approach. Also, $\soft(.)$ is the element-wise soft thresholding operator
\begin{equation} \label{eq:soft}
\soft_{\lambda}(z) = \sign(z)(|z| - \lambda)_+,
\end{equation}
where $(z)_+ = \text{max}(z,0)$. Finally,
\begin{equation} \label{eq:tvdn}
\TVDN_{\lambda}(Z) := \arg \min_X ~ \frac{1}{2} \|Z - X\|_F^2 + \lambda ~ \TV(X)
\end{equation}
is the solution to the (weighted) total variation denoising problem with data $Z$ (\ie, the proximal operator of $\TV(.)$). Problem \eqref{eq:tvdn} does not have a closed-form solution, so one must employ an iterative algorithm. To that end, we can equivalently express \eqref{eq:tvdn} as
\begin{equation} \label{eq:tvdn2}
\min_{x} ~ \frac{1}{2}\|z - x\|_2^2 + \lambda \|WCx\|_1,
\end{equation}
where $z = \vect(Z)$, $W$ is a diagonal matrix with $0/1$ entries encoding the indicator variables $w_{ijk}$, and $C$ is a sparse matrix that computes the first differences along each dimension of the (vectorized) video $x$. We solve \eqref{eq:tvdn2} via the alternating direction method of multipliers \cite{boyd2011distributed}, which prescribes the updates
\begin{equation} \label{eq:TV_admm_1}
\begin{aligned}
x^{k+1} &= \arg \min_{x} ~  \frac{1}{2} \| z - x \|^2 + \frac{\rho}{2} \| WCx - v^k + u^k \|_2^2 \\[2pt]
v^{k+1} &= \arg \min_{v} ~ \lambda \| v \|_1 + \frac{\rho}{2} \| WCx^{k+1} - v + u^k \|_2^2 \\[2pt]
u^{k+1} &=  u^k + WCx^{k+1} - v^{k+1},
\end{aligned}
\end{equation}
for some $\rho > 0$. In the static camera case---when $w_{ijk} \equiv 1$ in \eqref{eq:TV}---and circulant boundary conditions are assumed, one can efficiently compute the solution to the $x$ update in \eqref{eq:TV_admm_1} using fast Fourier transform operations. In the general case, the $x$ update is quadratic and can be computed via many off-the-shelf algorithms (e.g., conjugate gradient). The $v$-update is a simple soft thresholding operation, $v^{k+1} = \soft_{\lambda/\rho}(WCx^{k+1} + u^k)$. Algorithm~\ref{alg:proposed} summarizes the proposed algorithm.


\begin{algorithm}
\begin{algorithmic}[1]
\State \textbf{Input:} Video frames $F_1$, $\dots$, $F_p$
\State Compute registered frames $\widetilde{F}_1 \dots \widetilde{F}_p$ via \eqref{eq:registration}
\State Construct $Y$ and $M$ matrices via \eqref{eq:Ymatrix}
\State \textbf{Initialization:} $L^0 = U^0 = Y$, $S_1^0 = S_2^0 = 0$, and $k = 0$
\While {~not converged~}
\State  Update $L^k$, $S_1^k$, $S_2^k,$ and $U^k$ via \eqref{eq:rpca_apgupdates_optshrink}
\State $k = k + 1$
\EndWhile		
\State \textbf{Output:} Decomposition $\{L, \ S_1, \ S_2\}$
\end{algorithmic}
\caption{Proposed Algorithm}
\label{alg:proposed}
\end{algorithm}

\begin{figure*}[t!]
\centering
\centerline{\includegraphics[width=1\linewidth]{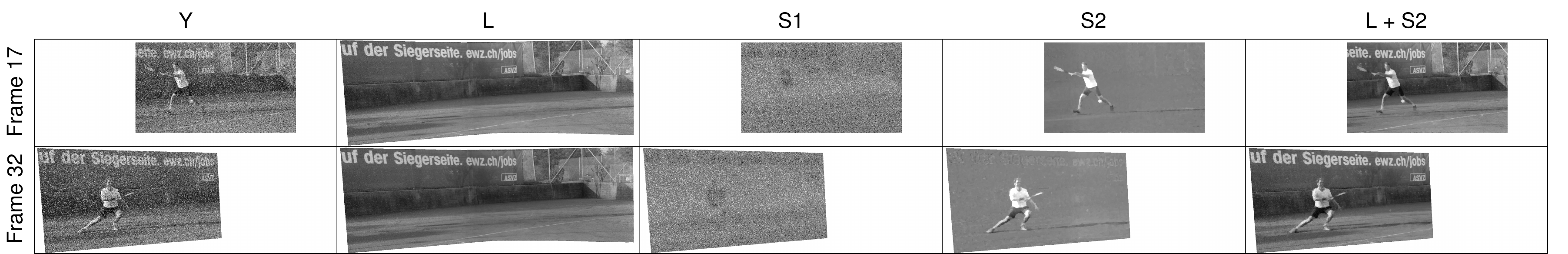}}
\caption{Proposed algorithm applied to the Tennis dataset corrupted by 30$\%$ outliers. $Y$: registered corrupted frames; $L$: reconstructed (panoramic) background; $S_1$: decoupled sparse corruptions; $S_2$: dynamic foreground; $L + S_2$: reconstructed scene.}
\label{fig:moving_camera}
\end{figure*}

\begin{table*}[t!]
\resizebox{\textwidth}{!}{
\begin{tabular}{|c|ccc|ccc|ccc|ccc|}
\hline
\multirow{2}{*}{Sequence} &
\multicolumn{3}{c|}{Proposed} &
\multicolumn{3}{c|}{RPCA} &
\multicolumn{3}{c|}{TVRPCA} &
\multicolumn{3}{c|}{DECOLOR} \\
\cline{2-13}
& f-PSNR & b-PSNR & F-measure & f-PSNR & b-PSNR & F-measure & f-PSNR & b-PSNR & F-measure & f-PSNR & b-PSNR & F-measure \\
\hline \hline	
Hall          & \textbf{38.94} & \textbf{37.98} & \textbf{0.60} & 27.12 & 32.63 & 0.19 & 36.50 &         37.42  & \textbf{0.60} & 27.02 & 31.63 & 0.17 \\
Fountain      & \textbf{39.73} & \textbf{35.48} & \textbf{0.74} & 26.99 & 32.06 & 0.21 & 36.87 & \textbf{35.48} &         0.72  & 26.89 & 30.69 & 0.15 \\
Escalator     & \textbf{33.15} & \textbf{31.56} & \textbf{0.72} & 23.45 & 26.27 & 0.35 & 30.91 &         30.96  &         0.69  & 23.27 & 22.17 & 0.25 \\
Water Surface & \textbf{42.14} & \textbf{36.96} & \textbf{0.94} & 22.92 & 31.45 & 0.40 & 40.14 &         36.81  &         0.82  & 22.12 & 20.66 & 0.26 \\
Shopping Mall & \textbf{40.26} &         39.83  & \textbf{0.74} & 25.06 & 34.62 & 0.31 & 37.43 & \textbf{40.88} &         0.73  & 25.01 & 31.42 & 0.26 \\
\hline \hline
Average       & \textbf{38.84} & \textbf{36.36} & \textbf{0.75} & 25.11 & 31.41 & 0.29 & 36.37 &         36.31  &         0.71  & 24.86 & 27.31 & 0.22 \\
\hline
\end{tabular}}
\caption{Performance metrics for each algorithm on sequences from the I2R dataset corrupted by 20\% outliers (salt and pepper).}
\label{tab:I2R_outliers}
\end{table*}

\begin{table*}[t!]
\resizebox{\textwidth}{!}{
\begin{tabular}{|c|ccc|ccc|ccc|ccc|}
\hline
\multirow{2}{*}{Sequence} &
\multicolumn{3}{c|}{Proposed} &
\multicolumn{3}{c|}{RPCA} &
\multicolumn{3}{c|}{TVRPCA} &
\multicolumn{3}{c|}{DECOLOR} \\
\cline{2-13}
& f-PSNR & b-PSNR & F-measure & f-PSNR & b-PSNR & F-measure & f-PSNR & b-PSNR & F-measure & f-PSNR & b-PSNR & F-measure \\
\hline \hline
Hall          & \textbf{36.66} & \textbf{32.72} &         0.58  & 31.80 & 30.14 & 0.30 & 34.64 & 21.83 & \textbf{0.59} & 31.65 & 25.14 &         0.56  \\
Fountain      & \textbf{38.14} & \textbf{30.05} & \textbf{0.74} & 34.57 & 29.35 & 0.35 & 36.45 & 24.22 &         0.70  & 36.51 & 25.54 &         0.71  \\
Escalator     & \textbf{32.83} & \textbf{26.60} & \textbf{0.72} & 29.87 & 25.07 & 0.49 & 31.15 & 22.35 &         0.68  & 25.67 & 23.54 & \textbf{0.72} \\
Water Surface & \textbf{38.46} & \textbf{31.08} & \textbf{0.94} & 30.19 & 28.71 & 0.57 & 33.83 & 23.88 &         0.81  & 29.35 & 20.88 &         0.84  \\
Shopping Mall & \textbf{37.31} & \textbf{35.29} & \textbf{0.71} & 32.34 & 31.54 & 0.34 & 35.13 & 24.31 & \textbf{0.71} & 32.39 & 30.93 & \textbf{0.71} \\
\hline \hline
Average       & \textbf{36.68} & \textbf{31.15} & \textbf{0.74} & 31.75 & 28.96 & 0.41 & 34.24 & 23.32 &         0.70  & 31.11 & 25.21 &         0.71  \\
\hline
\end{tabular}}
\caption{Performance metrics for each algorithm on sequences from the I2R dataset corrupted by 30dB Gaussian noise.}
\label{tab:I2R_noise}
\end{table*}

\begin{table*}[t!]
\resizebox{\textwidth}{!}{
\begin{tabular}{|c|ccc|ccc|ccc|ccc|}
\hline
\multirow{2}{*}{p} &
\multicolumn{3}{c|}{Proposed} &
\multicolumn{3}{c|}{RPCA} &
\multicolumn{3}{c|}{TVRPCA} &
\multicolumn{3}{c|}{DECOLOR} \\
\cline{2-13}
& f-PSNR & b-PSNR & F-measure & f-PSNR & b-PSNR & F-measure & f-PSNR & b-PSNR & F-measure & f-PSNR & b-PSNR & F-measure \\
\hline		
10\% & \textbf{41.48} & \textbf{39.37} & \textbf{0.60} & 30.35 & 32.67 & 0.27 & 38.38 & 38.98 & \textbf{0.60} & 30.28 & 31.54 & 0.29 \\		
20\% & \textbf{38.94} & \textbf{37.98} & \textbf{0.60} & 27.12 & 32.63 & 0.19 & 36.50 & 37.42 & \textbf{0.60} & 27.02 & 31.63 & 0.17 \\
30\% & \textbf{37.69} & \textbf{36.21} & \textbf{0.59} & 25.40 & 32.39 & 0.15 & 34.94 & 36.08 &         0.58  & 30.27 & 31.54 & 0.29 \\
40\% & \textbf{36.49} & \textbf{34.73} & \textbf{0.58} & 24.26 & 32.03 & 0.13 & 32.51 & 24.13 &         0.57  & 24.13 & 18.50 & 0.07 \\
50\% & \textbf{35.84} & \textbf{33.73} & \textbf{0.57} & 23.57 & 31.49 & 0.12 & 29.85 & 18.11 &         0.49  & 23.47 & 14.61 & 0.07 \\
60\% & \textbf{34.93} & \textbf{32.38} & \textbf{0.56} & 22.87 & 31.36 & 0.10 & 27.98 & 14.65 &         0.35  & 22.79 & 14.13 & 0.07 \\
\hline
\end{tabular}}
\caption{Performance metrics for each algorithm on the I2R Hall sequence as a function of (salt and pepper) outlier probability, p.}
\label{tab:Hall_outliers}
\end{table*}

\begin{table*}[t!]
\resizebox{\textwidth}{!}{
\begin{tabular}{|c|ccc|ccc|ccc|ccc|}
\hline
\multirow{2}{*}{SNR} &
\multicolumn{3}{c|}{Proposed} &
\multicolumn{3}{c|}{RPCA} &
\multicolumn{3}{c|}{TVRPCA} &
\multicolumn{3}{c|}{DECOLOR} \\
\cline{2-13}
& f-PSNR & b-PSNR & F-measure & f-PSNR & b-PSNR & F-measure & f-PSNR & b-PSNR & F-measure & f-PSNR & b-PSNR & F-measure \\
\hline		
 5dB & \textbf{31.78} & \textbf{26.15} & \textbf{0.52} & 20.85 & 18.55 & 0.07 & 25.20 & 11.29 &         0.08  & 27.98 & 14.30 &         0.07  \\
10dB & \textbf{32.78} & \textbf{27.87} & \textbf{0.54} & 23.04 & 23.31 & 0.08 & 26.85 & 13.33 &         0.14  & 28.54 & 14.30 &         0.07  \\
20dB & \textbf{34.73} & \textbf{30.73} & \textbf{0.56} & 27.42 & 28.73 & 0.14 & 30.20 & 16.89 &         0.34  & 30.13 & 14.30 &         0.07  \\
30dB & \textbf{36.66} & \textbf{32.72} &         0.58  & 31.80 & 30.14 & 0.30 & 34.64 & 21.83 & \textbf{0.59} & 31.65 & 25.14 &         0.56  \\
40dB & \textbf{39.64} & \textbf{33.90} & \textbf{0.60} & 36.20 & 31.27 & 0.46 & 37.96 & 25.70 &         0.58  & 36.27 & 31.51 &         0.59  \\
50dB & \textbf{42.89} & \textbf{36.14} &         0.60  & 40.59 & 32.00 & 0.54 & 41.47 & 29.77 &         0.59  & 37.87 & 32.73 & \textbf{0.61} \\
\hline
\end{tabular}}
\caption{Performance metrics for each algorithm on the I2R Hall sequence as a function of SNR (Gaussian noise).}
\label{tab:Hall_noise}
\vspace{-1mm}
\end{table*}

\section{Results} \label{sec:result}
To demonstrate the performance of our proposed algorithm, we first compare to the recent RPCA \cite{candes2011robust}, TVRPCA \cite{cao2016total}, and DECOLOR \cite{zhou2013moving} algorithms on corrupted static camera videos. We then demonstrate the ability of our algorithm to process corrupted moving camera videos, a scenario that the other methods cannot handle.

\subsection{Static camera} \label{subsec:static_camera}
We work with the I2R dataset\footnote{See http://perception.i2r.a-star.edu.sg/bk\_model/bk\_index.html.} of static camera sequences. Each sequence has between 523 and 3584 frames, each with a subset of 20 frames with labeled foreground masks. We select a subset of several hundred (contiguous) frames from each sequence containing 10 labeled frames.

To evaluate the denoising capabilities of each algorithm, we measure the peak signal-to-noise ratio of the foreground (f-PSNR) and background (b-PSNR) pixels, respectively, in decibels (dB) of the frames with labeled masks. We also measure the ability of each algorithm to isolate the foreground by applying a simple thresholding strategy to the foreground component of each algorithm ($S_2$ for our proposed method, $S$ for RPCA, and $F$ for TVRPCA) and then computing the F-measure of these estimated masks with respect to the labeled masks.\footnote{For DECOLOR, we use the foreground mask returned by the algorithm.} Here, F-measure is defined in terms of the precision and recall of the estimated mask as
\begin{equation} \label{eq:fmeasure}
F_{\text{measure}} = 2\times \dfrac{\text{Precision} \times \text{Recall}}{\text{Precision} + \text{Recall}},
\end{equation}
where $F_{\text{measure}} = 1$ corresponds to perfect accuracy. We tune the parameters of each algorithm for each metric and dataset individually.

Tables~\ref{tab:I2R_outliers} and \ref{tab:I2R_noise} compare the performance of the algorithms on I2R sequences corrupted by 20\% outliers (salt and pepper) and Gaussian noise with 30dB SNR, respectively. Tables~\ref{tab:Hall_outliers} and \ref{tab:Hall_noise} show the performance of each algorithm on the Hall sequence as a function of outlier probability and SNR, respectively. Our proposed method performs better than the other methods in most cases. 

The performance of RPCA and DECOLOR degrades dramatically when outliers are added because they lack the ability to separate outliers and other non-idealities from the dynamic foreground component. While TVRPCA performs better than these methods in the presence of outliers, our proposed method consistently achieves higher foreground PSNR and F-measure, which suggests that our algorithm can better decompose the scene into foreground and background components.

\subsection{Moving camera} \label{subsec:moving_camera}
Next, we demonstrate the performance of our proposed method on a moving camera sequence from a recent benchmark dataset~\cite{Perazzi2016}. The Tennis sequence consists of $35$ frames, each with resolution $480 \times 854$, of a camera panning across a tennis court with a player swinging a racket in the foreground.

Figure~\ref{fig:moving_camera} shows the outputs of Algorithm~\ref{alg:proposed} on the Tennis sequence corrupted by 30\% outliers (salt and pepper). The parameters used were $\tau = 0.33$, $r=1$, $\lambda_{S_1} = 0.001$, and $\lambda_{S_2} = 0.001$. Our proposed method gracefully aggregates the background information from the corrupted frames to produce a clean panoramic estimate ($L$) of the full field of view. Also, the TV-regularized component ($S_2$) is able to estimate the dynamic foreground (person) and decouple it from the sparse corruptions ($S_1$). None of the methods considered in Section~\ref{subsec:static_camera} can produce comparable results.


\section{Conclusion} \label{sec:conclusion}
\vspace{-1mm}
We proposed an augmented robust PCA algorithm for jointly estimating the foreground and background of a scene from noisy, moving camera video. Our proposed approach relies on a recently-developed low-rank matrix estimator (OptShrink) and weighted total variation regularization to recover the respective components of the scene. Our experimental results indicate that our algorithm is robust to both dense and sparse corruptions of the raw video and yields superior foreground-background separations compared to existing methods. In future work, we hope to investigate the usefulness of the foreground components produced by our algorithm for computer vision tasks like object tracking and activity detection.

\bibliographystyle{IEEEbib}
\bibliography{bibtex}

\begin{thebibliography}{10}

\bibitem{huang2011advanced}
S.-C. Huang,
\newblock ``An advanced motion detection algorithm with video quality analysis
  for video surveillance systems,''
\newblock {\em IEEE Transactions on Circuits and Systems for Video Technology},
  vol. 21, no. 1, pp. 1--14, 2011.

\bibitem{tsaig2002automatic}
Y.~Tsaig and A.~Averbuch,
\newblock ``Automatic segmentation of moving objects in video sequences: a
  region labeling approach,''
\newblock {\em IEEE Transactions on Circuits and Systems for Video Technology},
  vol. 12, no. 7, pp. 597--612, 2002.

\bibitem{bouwmans2014robust}
T.~Bouwmans and E.~H. Zahzah,
\newblock ``Robust pca via principal component pursuit: A review for a
  comparative evaluation in video surveillance,''
\newblock {\em Computer Vision and Image Understanding}, vol. 122, pp. 22--34,
  2014.

\bibitem{sobral2014comprehensive}
A.~Sobral and A.~Vacavant,
\newblock ``A comprehensive review of background subtraction algorithms
  evaluated with synthetic and real videos,''
\newblock {\em Computer Vision and Image Understanding}, vol. 122, pp. 4--21,
  2014.

\bibitem{ye2015foreground}
X.~Ye, J.~Yang, X.~Sun, K.~Li, C.~Hou, and Y.~Wang,
\newblock ``Foreground--background separation from video clips via
  motion-assisted matrix restoration,''
\newblock {\em IEEE Transactions on Circuits and Systems for Video Technology},
  vol. 25, no. 11, pp. 1721--1734, 2015.

\bibitem{he2012incremental}
J.~He, L.~Balzano, and A.~Szlam,
\newblock ``Incremental gradient on the grassmannian for online foreground and
  background separation in subsampled video,''
\newblock in {\em IEEE Conference on Computer Vision and Pattern Recognition},
  2012, pp. 1568--1575.

\bibitem{elgammal2000non}
A.~Elgammal, D.~Harwood, and L.~Davis,
\newblock ``Non-parametric model for background subtraction,''
\newblock in {\em European Conference on Computer Vision}, 2000, pp. 751--767.

\bibitem{candes2011robust}
E.~J. Cand{\`e}s, X.~Li, Y.~Ma, and J.~Wright,
\newblock ``Robust principal component analysis?,''
\newblock {\em Journal of the ACM}, vol. 58, no. 3, pp. 11, 2011.

\bibitem{guyon2012foreground}
C.~Guyon, T.~Bouwmans, and E.-H. Zahzah,
\newblock ``Foreground detection via robust low rank matrix decomposition
  including spatio-temporal constraint,''
\newblock in {\em Asian Conference on Computer Vision}, 2012, pp. 315--320.

\bibitem{zhou2013shifted}
T.~Zhou and D.~Tao,
\newblock ``Shifted subspaces tracking on sparse outlier for motion
  segmentation.,''
\newblock in {\em Artificial Intelligence Journal}, 2013.

\bibitem{zhang2014novel}
Z.~Zhang, G.~Ely, S.~Aeron, N.~Hao, and M.~Kilmer,
\newblock ``Novel methods for multilinear data completion and de-noising based
  on tensor-svd,''
\newblock in {\em IEEE Conference on Computer Vision and Pattern Recognition},
  2014, pp. 3842--3849.

\bibitem{stauffer1999adaptive}
C.~Stauffer and W.~E.~L. Grimson,
\newblock ``Adaptive background mixture models for real-time tracking,''
\newblock in {\em IEEE Conference on Computer Vision and Pattern Recognition},
  1999, vol.~2, pp. 246--252.

\bibitem{cao2016total}
X.~Cao, L.~Yang, and X.~Guo,
\newblock ``Total variation regularized rpca for irregularly moving object
  detection under dynamic background,''
\newblock {\em IEEE Transactions on Cybernetics}, vol. 46, no. 4, pp.
  1014--1027, 2016.

\bibitem{ebadi2016approximated}
S.~E. Ebadi, V.~G. Ones, and E.~Izquierdo,
\newblock ``Approximated robust principal component analysis for improved
  general scene background subtraction,''
\newblock {\em arXiv preprint arXiv:1603.05875}, 2016.

\bibitem{zhou2013moving}
X.~Zhou, C.~Yang, and W.~Yu,
\newblock ``Moving object detection by detecting contiguous outliers in the
  low-rank representation,''
\newblock {\em IEEE Transactions on Pattern Analysis and Machine Intelligence},
  vol. 35, no. 3, pp. 597--610, 2013.

\bibitem{forsyth2002computer}
D.~A. Forsyth and J.~Ponce,
\newblock {\em Computer Vision: A Modern Approach},
\newblock Prentice-Hall, 2002.

\bibitem{bay2006surf}
H.~Bay, T.~Tuytelaars, and L.~Van~Gool,
\newblock ``{SURF}: Speeded up robust features,''
\newblock in {\em European Conference on Computer Vision}, 2006, pp. 404--417.

\bibitem{fischler1981random}
M.~A. Fischler and R.~C. Bolles,
\newblock ``Random sample consensus: a paradigm for model fitting with
  applications to image analysis and automated cartography,''
\newblock {\em Communications of the ACM}, vol. 24, no. 6, pp. 381--395, 1981.

\bibitem{chan2011augmented}
S.~H. Chan, R.~Khoshabeh, K.~B. Gibson, P.~E. Gill, and T.~Q. Nguyen,
\newblock ``An augmented lagrangian method for total variation video
  restoration,''
\newblock {\em IEEE Transactions on Image Processing}, vol. 20, no. 11, pp.
  3097--3111, 2011.

\bibitem{parikh2014proximal}
N.~Parikh, S.~Boyd, et~al.,
\newblock ``Proximal algorithms,''
\newblock {\em Foundations and Trends in Optimization}, vol. 1, no. 3, pp.
  127--239, 2014.

\bibitem{cai2010singular}
J.-F. Cai, E.~J. Cand{\`e}s, and Z.~Shen,
\newblock ``A singular value thresholding algorithm for matrix completion,''
\newblock {\em SIAM Journal on Optimization}, vol. 20, no. 4, pp. 1956--1982,
  2010.

\bibitem{moore2014improved}
B.~E. Moore, R.~R. Nadakuditi, and J.~A. Fessler,
\newblock ``Improved robust pca using low-rank denoising with optimal singular
  value shrinkage,''
\newblock in {\em IEEE Workshop on Statistical Signal Processing}, 2014, pp.
  13--16.

\bibitem{nadakuditi2014optshrink}
R.~R. Nadakuditi,
\newblock ``Optshrink: An algorithm for improved low-rank signal matrix
  denoising by optimal, data-driven singular value shrinkage,''
\newblock {\em IEEE Transactions on Information Theory}, vol. 60, no. 5, pp.
  3002--3018, 2014.

\bibitem{boyd2011distributed}
S.~Boyd, N.~Parikh, E.~Chu, B.~Peleato, and J.~Eckstein,
\newblock ``Distributed optimization and statistical learning via the
  alternating direction method of multipliers,''
\newblock {\em Foundations and Trends{\textregistered} in Machine Learning},
  vol. 3, no. 1, pp. 1--122, 2011.

\bibitem{Perazzi2016}
F.~Perazzi, J.~Pont-Tuset, B.~McWilliams, L.~{Van Gool}, M.~Gross, and
  A.~Sorkine-Hornung,
\newblock ``A benchmark dataset and evaluation methodology for video object
  segmentation,''
\newblock in {\em IEEE Conference on Computer Vision and Pattern Recognition},
  2016.

\end{thebibliography}

\end{document}